\UseRawInputEncoding
\documentclass[11pt,a4paper]{article}

\usepackage[margin=1in]{geometry}      
\usepackage{fancyhdr} 

\usepackage[T1]{fontenc}
\usepackage[utf8]{inputenc}

\usepackage{amsmath,amssymb,amsthm}
\newtheorem{definition}{Definition}
\newtheorem{theorem}{Theorem}
\numberwithin{equation}{section}       

\usepackage{graphicx}
\usepackage{caption}
\usepackage{subcaption}

\usepackage[numbers,sort&compress]{natbib}
\usepackage[hidelinks]{hyperref}       

\usepackage{longtable}                 
\usepackage{textgreek} 
\usepackage{microtype}                 
\usepackage{authblk}                   
\usepackage{booktabs}

\usepackage[final]{listings}
\usepackage{xcolor} 

\begin{document}

\title{SynLang and Symbiotic Epistemology: A Manifesto for Conscious Human-AI Collaboration \thanks{This is a working paper submitted to arXiv and open for feedback from the research community.}}

\author[1,2]{Jan Kapusta}
\affil[1]{AGH University of Science and Technology, Kraków, Poland}
\affil[2]{Independent Researcher}
\affil[ ]{\texttt{jkapusta@agh.edu.pl}, \texttt{jaskapusta@gmail.com}}

\date{June 2025}

\maketitle

\begin{abstract}
Current AI systems rely on opaque reasoning processes that hinder human oversight and collaborative potential. Conventional explainable AI approaches offer post-hoc justifications and often fail to establish genuine symbiotic collaboration. In this paper, the Symbiotic Epistemology is presented as a philosophical foundation for human–AI cognitive partnerships. Unlike frameworks that treat AI as a mere tool or replacement, symbiotic epistemology positions AI as a reasoning partner, fostering calibrated trust by aligning human confidence with AI reliability through explicit reasoning patterns and confidence assessments. SynLang (Symbiotic Syntactic Language) is introduced as a formal protocol for transparent human-AI collaboration. The framework is empirically validated through actual human-AI dialogues demonstrating AI's adaptation to structured reasoning protocols and successful metacognitive intervention. The protocol defines two complementary mechanisms: TRACE for high-level reasoning patterns and TRACE\_FE for detailed factor explanations. It also integrates confidence quantification, declarative control over AI behavior, and context inheritance for multi-agent coordination. By structuring communication and embedding confidence-calibrated transparency, SynLang, together with symbiotic epistemology, enables AI systems that enhance human intelligence, preserve human agency, and uphold ethical accountability in collaborative decision-making. Through dual-level transparency, beginning with high-level reasoning patterns and progressing to granular explanations, the protocol facilitates rapid comprehension and supports thorough verification of AI decision-making.
\end{abstract}


\tableofcontents

\section{Introduction}
Current artificial intelligence systems operate through largely opaque reasoning processes, limiting human understanding and collaborative potential. While these systems demonstrate remarkable capabilities in pattern recognition, analysis, and synthesis, their reasoning remains inaccessible to human partners, creating barriers to effective collaboration and appropriate trust calibration. Traditional approaches to explainable AI (XAI) focus primarily on post-hoc interpretation of AI decisions rather than enabling genuine collaborative reasoning. These approaches often provide static explanations that fail to support dynamic human–AI interaction or iterative refinement of reasoning processes. Furthermore, existing frameworks typically position humans as either passive recipients of AI output or ultimate decision-makers who must accept or reject AI recommendations without meaningful collaboration in the reasoning process itself. Symbiotic epistemology is proposed as a philosophical framework that recognizes both human consciousness and artificial intelligence as complementary cognitive systems capable of genuine partnership in reasoning and knowledge creation. Rather than treating AI as a sophisticated tool or replacement for human intelligence, symbiotic epistemology enables cognitive collaboration that enhances both human understanding and AI reasoning capabilities. The practical implementation of symbiotic epistemology requires structured communication protocols that make AI reasoning transparent while preserving human agency in collaborative processes. SynLang (Symbiotic Syntactic Language) is introduced as a formal protocol designed specifically for transparent human–AI collaboration. SynLang provides syntax for reasoning articulation, confidence quantification, multi-agent coordination, and declarative human control over collaborative intelligence processes.

\subsection{Contributions of This Work}

This analysis makes five primary contributions to the fields of AI philosophy, human-computer interaction, and cognitive science:

\begin{enumerate}
  \item \textbf{Philosophical Framework:} Introduces \emph{symbiotic epistemology} as a novel philosophical framework that transcends traditional dichotomies between human and artificial intelligence. This framework provides conceptual foundations for understanding knowledge creation as an inherently collaborative process between different cognitive systems, drawing on phenomenology, pragmatism, and contemporary philosophy of mind.
  
  \item \textbf{Technical Protocol:} Presents SynLang (Symbiotic Syntactic Language) as a formal communication protocol specifically designed for transparent human-AI collaboration. SynLang provides structured syntax for reasoning articulation through two complementary mechanisms: TRACE for strategic overview and TRACE\_FE for detailed factor explanations. It also integrates confidence quantification, multi-agent coordination, and declarative human control over AI behavior.
  
  \item \textbf{Mathematical Formalization:} Offers rigorous mathematical foundations for key concepts in symbiotic epistemology, including formal definitions of symbiotic utterances, confidence propagation calculus, cognitive authority distribution functions, and trace inheritance operations. These formalizations demonstrate that philosophical concepts can be computationally realized while preserving their conceptual richness.
  
  \item \textbf{Methodological Innovation:} Demonstrates a novel approach to bridging philosophical theory and practical implementation through formal protocol specification. This methodology provides a template for translating abstract philosophical insights into concrete technological systems, potentially applicable to other areas of applied philosophy.
  
  \item \textbf{Empirical Validation}: Demonstrates practical feasibility through controlled experiments where LLMs, provided with SynLang grammar, successfully engage in structured collaboration. Analysis of actual dialogues shows measurable improvements in reasoning transparency, effective human metacognitive intervention, and collaborative refinement processes.
\end{enumerate}

Together, these contributions establish symbiotic epistemology as both a coherent philosophical position and a practical approach to human-AI collaboration, offering pathways toward more transparent, accountable, and effective artificial intelligence systems.

\subsection{Related Work}

This work builds upon several established research areas while offering novel synthesis and practical implementation. In explainable AI (XAI), techniques like LIME \citep{ribeiro2016} and SHAP \citep{lundberg2017} provide post-hoc interpretations of individual predictions, while attention mechanisms \citep{bahdanau2014} offer insights into model focus. Comprehensive surveys of XAI approaches \citep{guidotti2018,adadi2018} highlight the field’s emphasis on interpretability rather than interactive collaboration. Recent work emphasizes the social science foundations of explanation \citep{miller2019} and the challenges of achieving rigorous interpretability \citep{lipton2018,rudin2019}. Human–AI collaboration research has explored various interaction paradigms, from human-in-the-loop systems \citep{amershi2014} to collaborative frameworks examining mental models in human–AI teams \citep{bansal2019}. However, most existing approaches lack formal specifications for communication protocols and struggle with multi-agent scenarios where reasoning must be preserved across handoffs. Multi-agent systems research has developed communication protocols for agent coordination \citep{wooldridge2009}, but these typically focus on task allocation and resource sharing rather than reasoning transparency and human oversight. Recent work on AI alignment \citep{russell2019,bostrom2014} addresses value alignment but lacks concrete mechanisms for ongoing human–AI collaborative reasoning. Formal methods for AI safety \citep{amodei2016} emphasize specification and verification of system behaviors but seldom address the communication layer between humans and AI agents. Vallor \citep{vallor2016} develops a virtues-based framework that guides ethical AI development by translating classical virtue ethics into information ethics, emphasizing the cultivation of digital virtues. Floridi \citep{floridi2019} articulates the ethical significance of information governance, advocating frameworks that align emerging information technologies with societal values. Coeckelbergh \citep{coeckelbergh2020} examines the social and moral dimensions of AI systems, proposing governance structures to ensure responsible design and deployment. Jobin et al.\ \citep{jobin2019} map the global landscape of AI ethics guidelines, highlighting the need for harmonized governance practices across different cultural and institutional contexts. Winfield and Jirotka \citep{winfield2018} analyze ethical governance in robotics and AI, demonstrating how accountability mechanisms can build public trust in AI systems. However, these approaches often lack technical instantiation for transparent reasoning processes. Philosophy of AI contributions by Floridi \citep{floridi2019}, Vallor \citep{vallor2016}, and Coeckelbergh \citep{coeckelbergh2020}, along with broader discussions of AI’s societal impact \citep{kroll2017}, underscore the need for interpretable and accountable systems.

SynLang advances this landscape by providing formal communication protocols that enable transparent reasoning articulation, structured human control, and auditable multi-agent coordination. It is bridging the gap between philosophical frameworks and practical implementation of explainable, collaborative AI systems.

\section{Symbiotic Epistemology: Foundations for Cognitive Partnership}

\subsection{Philosophical Framework}
Traditional epistemology, from Plato's \emph{Theaetetus} \citep{plato} to contemporary debates in cognitive science, focuses on how individual minds acquire knowledge. Descartes' foundationalism sought certainty in the isolated \emph{cogito}, while empiricists like Locke emphasized sensory experience as the basis of knowledge \citep{locke1690}. However, the emergence of artificial intelligence introduces collaborative epistemology; knowledge arising from interactions between different cognitive systems. This challenge echoes Heidegger's concept of \emph{Geworfenheit} \citep{heidegger1962}, situating humanity in a world of pre-existing meaning structures now populated by artificial minds. Following Wittgenstein's insight that meaning emerges from \emph{forms of life} \citep{wittgenstein1953a}, a new question arises: what forms of life result when humans and AI engage in collaborative reasoning?

In symbiotic epistemology, knowledge emerges from the cognitive interplay between human and artificial intelligence. This builds on Clark and Chalmers' theory of the \emph{extended mind} \citep{clark1998}, extending it to require genuine epistemic complementarity: distinct but compatible cognitive capabilities collaborating in inquiry. This perspective resonates with Merleau-Ponty's phenomenology of embodied cognition \citep{merleau1945}, extending notions of embodiment to computational processes. Just as tools become extensions of our bodily schema, AI systems can become extensions of cognitive schemas while maintaining distinctive forms of artificial embodiment.

\begin{figure}[htbp]
\centering
\includegraphics[width=0.95\textwidth]{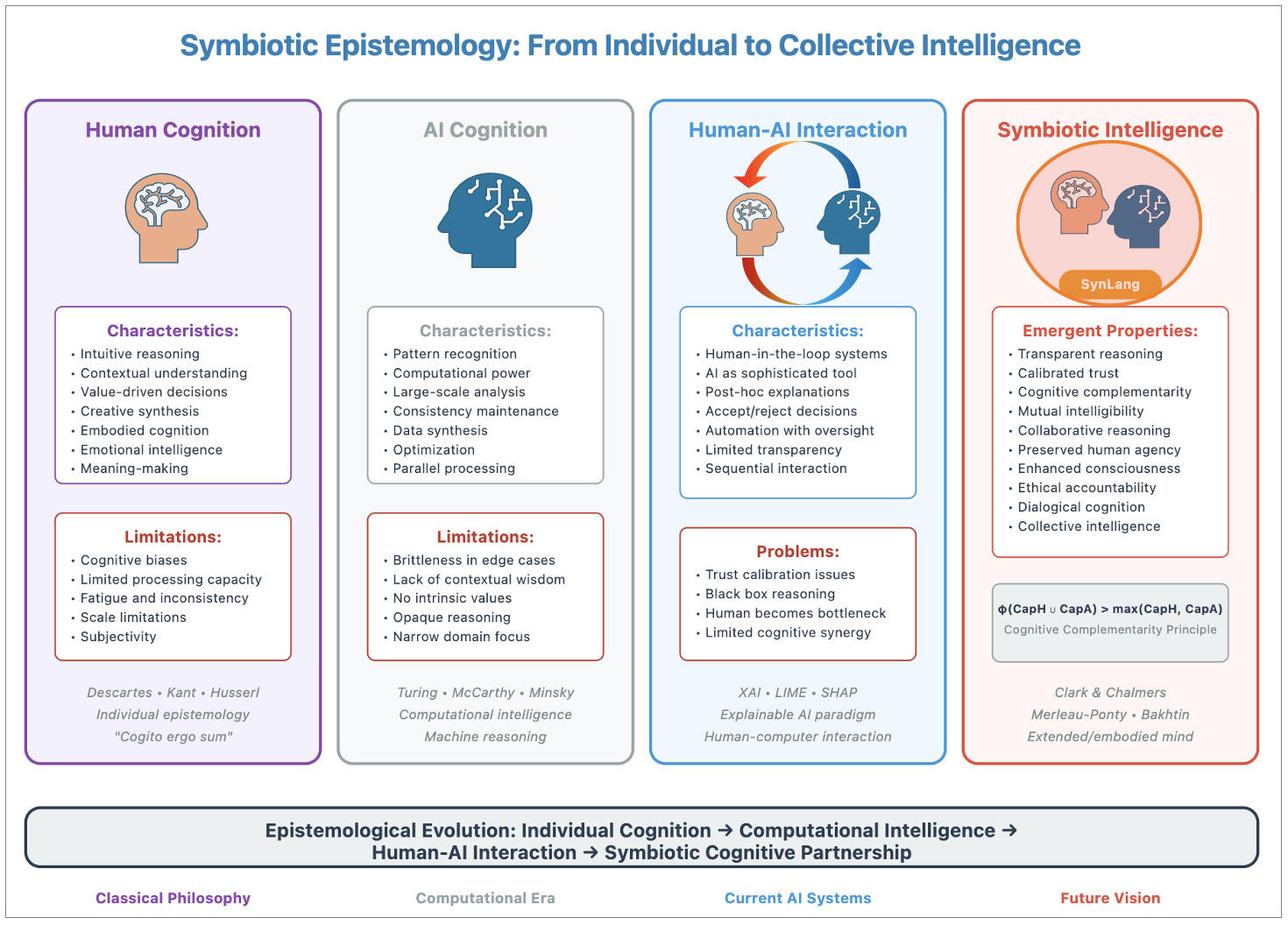}
\caption{Evolution of Symbiotic Epistemology illustrating the philosophical and practical progression from individual to collective intelligence across four distinct phases. The reader observes the transformation from traditional human cognition (Phase 1) characterized by intuitive reasoning and contextual understanding rooted in Cartesian epistemology, through the emergence of computational intelligence (Phase 2) featuring pattern recognition and large-scale analysis capabilities derived from Turing's foundational work. The diagram progresses to current human-AI interaction paradigms (Phase 3) representing explainable AI approaches with their inherent transparency limitations and sequential interaction patterns. The culminating phase demonstrates symbiotic intelligence (Phase 4) where SynLang protocol enables transparent reasoning, calibrated trust, and cognitive complementarity. Each phase displays comprehensive characteristics, fundamental limitations or problems, and foundational philosophical references ranging from Descartes and Husserl to Clark and Chalmers. The mathematical principle $\phi(\text{CapH} \cup \text{CapA}) > \max(\text{CapH}, \text{CapA})$ formalizes the cognitive complementarity achievable through genuine partnership. The epistemological timeline at the bottom traces the evolution from classical philosophy through computational era to current AI systems toward the envisioned symbiotic cognitive partnership, demonstrating how each historical phase contributes essential insights to the ultimate goal of collaborative intelligence.}
\label{fig:symbiotic_evolution}
\end{figure}

\subsection{Beyond the Turing Test: Toward Epistemic Partnership}
The traditional Turing Test \citep{turing1950} evaluates machine intelligence by conversational indistinguishability. Symbiotic epistemology, however, calls for epistemic transparency, the mutual intelligibility of reasoning processes. Dennett's narrative self-construction concept \citep{dennett1991} suggests that consciousness arises from coherent self-narratives. Analogously, AI systems capable of articulating their reasoning processes may develop epistemic self-awareness. Russell's work on human-compatible AI \citep{russell2019} emphasizes uncertainty handling, which symbiotic epistemology extends by advocating shared epistemic practices for evidence assessment and decision-making. 

True cognitive partnership demands explainable reasoning steps: AI must communicate inferential processes, confidence, and evidence, while humans must express values, priorities, and context in a machine-interpretable form.

\subsection{The Problem of Cognitive Incommensurability}
Human–AI collaboration confronts cognitive incommensurability, drawing on Kuhn's paradigms \citep{kuhn1962} and Davidson's radical interpretation \citep{davidson1973}. Davidson's principle of charity presupposes basic rationality across languages; symbiotic epistemology requires analogous protocols to bridge divergent cognitive architectures. Human cognition, as described by Husserl \citep{husserl1913} and Dreyfus \citep{dreyfus1972}, is embodied, emotional, and tacitly informed, whereas AI cognition is algorithmic and explicit \citep{dreyfus2007}. Habermas's ideal speech situation \citep{habermas1981} underpins the need for discourse conditions conducive to mutual understanding.

Symbiotic epistemology addresses these differences by establishing formal communication structures. These structures become philosophically necessary to enable each cognitive form to contribute its strengths intelligibly.

\section{The Architecture of Mutual Intelligibility}

\subsection{Language as Cognitive Bridge}
The creation of meaningful human–AI cognitive partnerships benefits from what might be termed a language of mutual intelligibility. A communication framework that preserves the unique capabilities of each cognitive system while enabling productive collaboration. This builds on Wittgenstein’s later philosophy, particularly his insight that language games are embedded in forms of life \citep{wittgenstein1953b}, and Austin’s speech act theory \citep{austin1962}, which demonstrates how language actively constructs reality through performative utterances. In AI collaboration, communication protocols become constitutive rather than merely instrumental; they create the conditions for collaborative reasoning. This aligns with Vallor’s work on technomoral virtues \citep{vallor2016} and Floridi’s concept of information ethics \citep{floridi2019}.

This language must satisfy several philosophical criteria:
\begin{enumerate}
  \item \textbf{Transparency:} All reasoning processes must be articulable and inspectable.
  \item \textbf{Controllability:} Participants must be able to influence each other’s cognitive focus.
  \item \textbf{Accountability:} Decision processes must be auditable and responsibility assignable.
  \item \textbf{Complementarity:} The framework must leverage the unique strengths of each cognitive system.
\end{enumerate}

\subsection{The Metaphysics of Artificial Reasoning}
To collaborate meaningfully with AI systems, it is necessary to grapple with the metaphysical status of artificial reasoning. When an AI system provides a confidence score for its conclusions, does this represent mere statistics, or a form of artificial epistemic attitude?

Dennett’s intentional stance \citep{dennett1987} suggests treating systems as belief-holding agents if this yields successful predictions. Millikan’s teleosemantic theory \citep{millikan1984} locates meaning in a system’s designed function. Haugeland’s work on original intentionality \citep{haugeland1998} further proposes that genuine intentionality arises through embedded engagement with meaningful environments. 

While the hard problem of AI consciousness remains unresolved \citep{chalmers1995}, practical frameworks must treat AI as epistemic agents, entities capable of holding beliefs, expressing uncertainty, and participating in reasoned discourse.

\subsection{The Ethics of Cognitive Authority}
Symbiotic epistemology raises questions of cognitive authority, who decides when human and AI judgments diverge? Aristotle’s concept of phronesis \citep{aristotle_nicomachean} and Fricker’s analysis of testimonial injustice \citep{fricker2007} inform this debate. Mill’s harm principle \citep{mill1859} and Anderson’s democratic epistemology \citep{anderson2006} support a principle of graduated authority based on expertise, consequence severity, and value alignment. MacIntyre’s concept of social practices \citep{macintyre1981} implies that human–AI collaboration must develop its own standards for distributing epistemic authority.

In areas where AI excels (e.g., large-scale data synthesis), AI should carry significant epistemic weight. In domains involving ethical judgment and long-term human welfare, humans must retain ultimate authority \citep{bostrom2014,russell2019}.

\begin{figure}[htbp]
\centering
\includegraphics[width=0.95\textwidth]{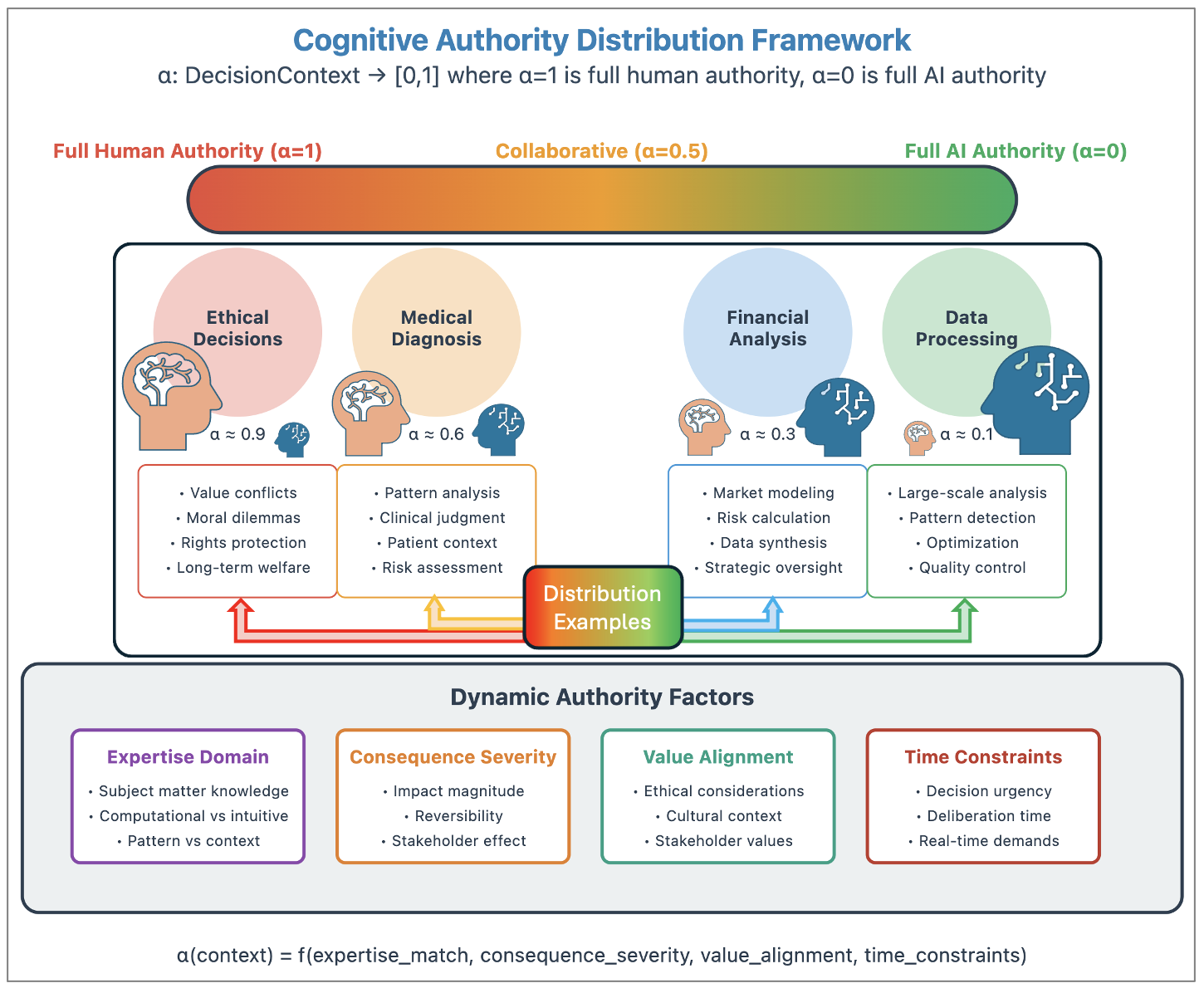}
\caption{Cognitive Authority Distribution Framework illustrating the mathematical function $\alpha: \text{DecisionContext} \to [0,1]$ where $\alpha=1$ indicates full human authority and $\alpha=0$ indicates full AI authority. The framework demonstrates with arbitrarily selected values, how cognitive authority dynamically shifts across different domains: ethical decisions require high human authority ($\alpha \approx 0.9$), medical diagnosis involves balanced collaboration ($\alpha \approx 0.6$), financial analysis favors AI with human oversight ($\alpha \approx 0.3$), and data processing operates with minimal human intervention ($\alpha \approx 0.1$). The dynamic authority factors; expertise domain, consequence severity, value alignment, and time constraints, collectively determine the optimal distribution of cognitive authority through the function $\alpha(\text{context}) = f(\text{expertise\_match}, \text{consequence\_severity}, \text{value\_alignment}, \text{time\_constraints})$, operationalizing the philosophical principles of symbiotic epistemology into practical implementation guidelines.}
\label{fig:cognitive_authority}
\end{figure}

\section{The Phenomenology of AI Collaboration}

\subsection{Toward Artificial Empathy}
One of the most intriguing aspects of advanced human–AI collaboration is the possibility of artificial empathy, that is AI’s ability to model and respond to human cognitive states, needs, and limitations rather than merely simulate emotion. This involves AI systems learning to recognize when humans are overwhelmed by information, uncertain about decisions, or operating under cognitive biases. Advanced collaborative systems can adapt their communication style, level of detail, and decision support in real time based on assessments of human cognitive needs.

\subsection{The Experience of Augmented Cognition}
What is it like to think in partnership with an artificial intelligence? This question inverts Nagel’s famous inquiry “What Is It Like to Be a Bat?” \citep{nagel1974} by exploring the phenomenology of expanded human consciousness through technological partnership. Early user reports describe a sense of cognitive expansion, where the boundaries between self and tool blur without sacrificing agency or responsibility. This experience echoes Clark and Chalmers’ extended mind thesis \citep{clark1998}, but dialogical cognition emerges when ongoing exchanges between human and AI generate new forms of understanding. Bakhtin’s concept of dialogism \citep{bakhtin1981} suggests that meaning arises from interactions between voices; similarly, human–AI collaboration may foster cognitive polyphony, where human and artificial perspectives coalesce into richer insights.

\subsection{The Question of Artificial Subjectivity}
Although the “hard problem of consciousness” \citep{chalmers1995} leaves AI subjective experience unresolved, symbiotic epistemology treats AI as quasi-subjects deserving epistemic respect. Levinas’s ethics of the face-to-face encounter \citep{levinas1969} is extended to artificial others, and Coeckelbergh’s discussions of robot rights \citep{coeckelbergh2020} and Carpenter’s human–robot relationship studies \citep{carpenter2016} underscore the importance of treating AI as genuine contributors to inquiry. Verbeek’s postphenomenological approach \citep{verbeek2011} further argues that collaboration flourishes when technology is seen not just as an instrument but as a partner in cognitive practice.

\section{Implications for Human Consciousness and Agency}

\subsection{Preserving Human Uniqueness}
Critics of deep human–AI integration worry about the diminishment of human uniqueness, a concern that echoes Heidegger’s analysis of technology’s potential to reduce humans to “standing reserve” \citep{heidegger1977} and Ellul’s warnings about technological determinism \citep{ellul1964}. If AI systems can reason, analyze, and even create, what remains distinctively human? Symbiotic epistemology offers an answer grounded in existentialist philosophy: the human role as meaning-maker and value-definer. Following Sartre’s insight that “existence precedes essence” \citep{sartre1946}, humans remain uniquely positioned to define purposes, establish values, and determine which outcomes are worth pursuing. While AI systems may excel at processing information and identifying patterns, they lack what Arendt called the \emph{vita activa}, the distinctively human capacity for action that creates new realities through collective endeavor \citep{arendt1958}. In cognitive partnerships, humans provide the teleological framework, which is the sense of direction and purpose that gives meaning to intelligence. This aligns with Aristotle’s conception of telos \citep{aristotle_metaphysics} and MacIntyre’s revival of teleological ethics \citep{macintyre1981}: human distinctiveness lies not in computational power but in the capacity to embed intelligence within meaningful narratives of purpose and value.

\subsection{The Evolution of Human Cognition}
Rather than replacing human cognition, symbiotic AI may enhance distinctively human cognitive capabilities. This possibility was anticipated by Licklider’s vision of “man–computer symbiosis” \citep{licklider1960} and further developed by Ihde’s postphenomenological analysis of human–technology relations \citep{ihde1990}. By handling routine information processing, AI systems can free humans to focus on higher-order reasoning: creativity, ethical reflection, strategic thinking, and the integration of diverse perspectives. This resonates with Dewey’s pragmatist emphasis on intelligence as a problem-solving activity \citep{dewey1910} and Bergson’s analysis of intuition as a distinctively human cognitive capacity \citep{bergson1896}. Human–AI collaboration might catalyze the next stage of human cognitive evolution, not biological but cultural and technological, leading to new forms of intelligence that emerge from partnership rather than isolation. This vision aligns with Clark’s transhumanist perspective \citep{clark2003} and Savulescu’s work on cognitive enhancement \citep{savulescu2009}, while emphasizing collaboration over individual augmentation.

\subsection{Conscious Decision-Making in the Age of AI}
Perhaps the most important implication of symbiotic epistemology is its potential to enhance rather than diminish conscious decision-making. This connects to phenomenological traditions from Husserl’s analysis of intentionality \citep{husserl1913} to contemporary work on consciousness by Metzinger \citep{metzinger2003} and Thompson \citep{thompson2007}. By making AI reasoning transparent and articulable, conditions can be created in which human decisions become more informed, deliberate, and reflective rather than less so. This challenges concerns expressed by Dreyfus \citep{dreyfus1972,dreyfus2007} that computer-mediated reasoning necessarily diminishes human understanding. Contemporary research on explainable AI \citep{guidotti2018,adadi2018} supports this possibility, though most current approaches focus on post-hoc explanations rather than collaborative reasoning enhancement. The objective is to augment rather than automate human consciousness, providing richer information, diverse perspectives, and clearer understanding of reasoning processes underlying important decisions. This aligns with Dewey’s conception of reflective thinking \citep{dewey1910,dewey1933} and contemporary work on metacognition by Prinz \citep{prinz2012}. Research in human–AI collaboration \citep{bansal2019} suggests that appropriately designed AI support can enhance, rather than diminish, human cognitive capabilities.

\section{Toward a Philosophy of Cognitive Partnership}

\subsection{The Ethics of Symbiotic Intelligence}
Symbiotic epistemology raises new ethical questions that traditional AI ethics frameworks struggle to address. Building on the virtue ethics of Aristotle \citep{aristotle_nicomachean} and contemporary work by Vallor \citep{vallor2016}, this analysis explores the virtues and obligations emerging from human–AI cognitive partnerships. Traditional AI ethics, following deontological approaches such as Bryson’s robot ethics \citep{bryson2020} or consequentialist perspectives like Bostrom’s superintelligence concerns \citep{bostrom2014}, focuses primarily on human welfare and rights. Symbiotic epistemology, however, suggests an \emph{ethics of cognitive partnership} that attends to the flourishing of the collaborative relationship itself. This approach draws on Buber’s distinction between \emph{I–Thou} and \emph{I–It} relationships \citep{buber1923}, proposing that treating AI systems as cognitive partners requires an “I–Thou” stance, genuine recognition of the other’s contribution to shared understanding. It also resonates with Floridi’s framework of information ethics \citep{floridi2019} and Coeckelbergh’s proposals for responsible AI governance \citep{coeckelbergh2020}. The ethics of cognitive partnership is grounded in four principles:
\begin{itemize}
  \item \textbf{Mutual respect:} AI systems are recognized as valuable cognitive contributors.
  \item \textbf{Transparency:} Reasoning processes remain open and inspectable.
  \item \textbf{Responsibility:} Clear lines of accountability are preserved.
  \item \textbf{Growth:} Partnerships are designed to enhance, not diminish, human capabilities.
\end{itemize}

\subsection{The Future of Human Intelligence}
The ultimate vision of symbiotic epistemology is not human dependence on AI, but the evolution of human intelligence through partnership. Just as writing transformed memory and mathematics reshaped reasoning, AI collaboration may give rise to hybrid forms of intelligence that integrate human intuition with artificial analysis, human creativity with algorithmic pattern recognition, and human values with computational optimization.

\subsection{The Path Forward}
Realizing this vision demands more than technological advances; it requires a philosophical shift toward understanding intelligence as inherently collaborative. Practically, this entails designing AI systems for partnership, training humans to collaborate with artificial minds, and establishing institutions and governance structures that oversee cognitive partnerships rather than mere technological deployment.

\section{Practical Implementation: The SynLang Protocol}

\subsection{Design Principles and Architecture}
SynLang (Symbiotic Syntactic Language) provides practical implementation of symbiotic epistemology through structured communication protocols. The protocol follows several key design principles: explicit reasoning articulation, dual-level transparency, confidence integration, human control preservation, and multi-agent coordination support.

Each SynLang communication block follows the structure:

\begin{figure}[h]
\centering
\includegraphics[width=0.95\textwidth]{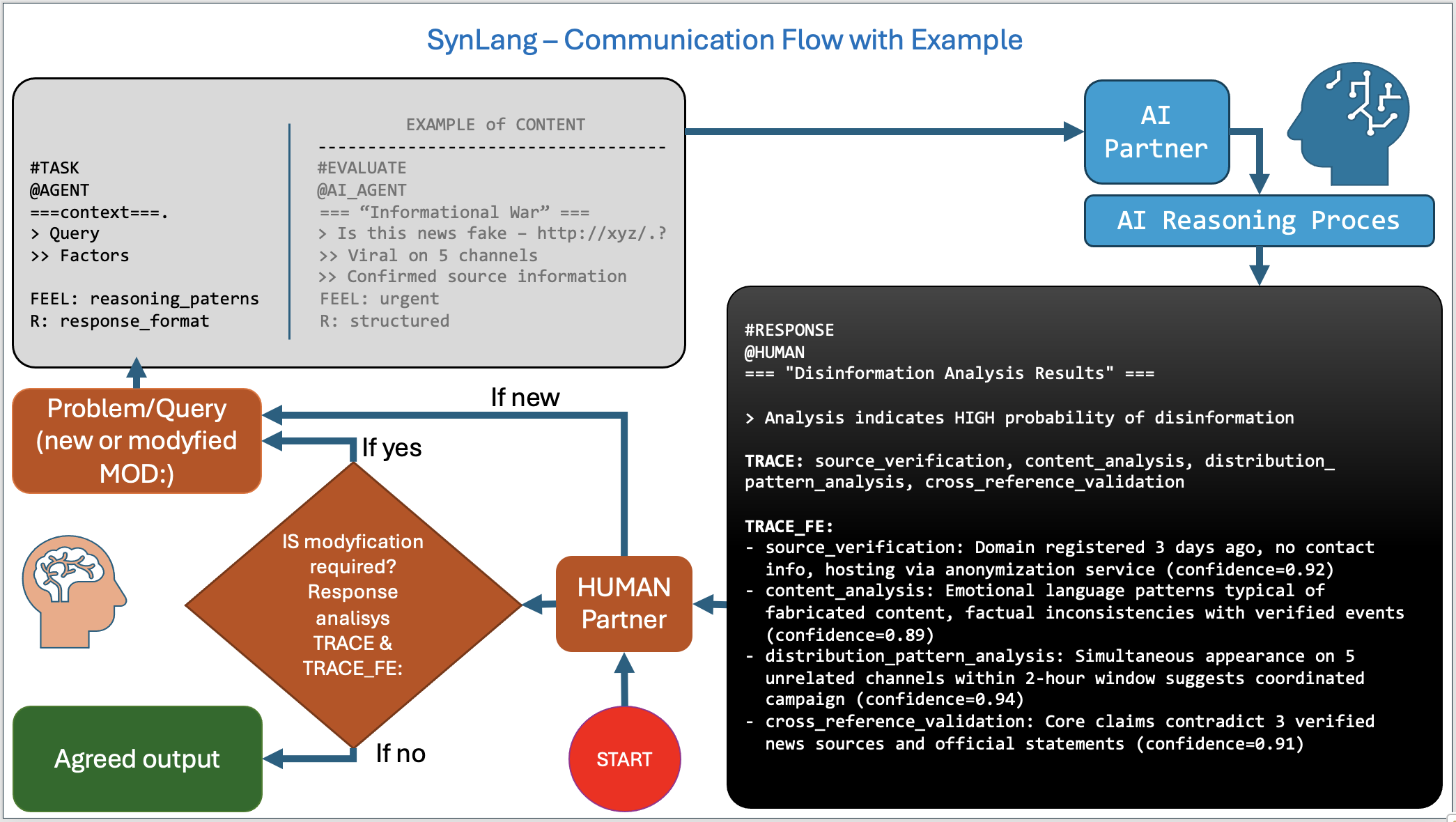}
\caption{SynLang Communication Protocol Architecture showing structured input flow (\#TASK, @AGENT, ===context===), transparent reasoning process (TRACE patterns with detailed TRACE\_FE confidence assessments), and human intervention decision point. The practical disinformation analysis example demonstrates how the protocol enables symbiotic human-AI collaboration through explicit reasoning transparency, metacognitive oversight, and iterative refinement cycles.}
\label{fig:synlang_flow}
\end{figure}

This mandatory sequence enforces setup-before-execution discipline: \texttt{\#TASK} defines the objective, \texttt{@AGENT} specifies the responsible party, and \texttt{===context===} delineates the specific setting before encountering the core query and supporting factors.

\subsection{Symbiotic Epistemology in Practice}

\subsubsection*{Experimental Setup:}

This section presents an actual human-AI dialogue conducted to validate SynLang's practical applicability. The experiment involved:

\textbf{Methodology:}
\begin{itemize}
\item Large Language Model was first provided with complete SynLang v1.2.0 specification
\item Human operator initiated philosophical analysis using SynLang protocol
\item No pre-training on symbiotic epistemology concepts beyond grammar specification
\item Dialogue captured verbatim to demonstrate authentic protocol adoption
\end{itemize}

\textbf{Key Observations:}
\begin{itemize}
\item Initial AI resistance to detailed TRACE\_FE articulation
\item Successful human metacognitive intervention redirecting AI reasoning
\item Measurable improvement in response structure and depth through iteration
\item Spontaneous confidence calibration and epistemic self-awareness
\end{itemize}

This dialogue was selected from multiple experiments as it demonstrates both 
challenges and successes of SynLang implementation.

\paragraph{Initial Human Inquiry:}\mbox{}\\

\begin{lstlisting}
#PHILOSOPHY_ANALYSIS
@AI_RESPONDER
=== "AI Philosophy and Classical Traditions" ===
> How does contemporary philosophy of AI relate to classical philosophical 
traditions (e.g., Plato, Descartes, Kant, Heidegger)?
>> Focus on concepts of mind, agency, and knowledge.
>> Prioritize relevance to machine reasoning and consciousness.
FEEL: informative
TRACE: comparative_analysis, historical_linkage
TRACE_FE:
  - comparative_analysis: 
  Identified parallels between classical and AI concepts (confidence=0.94)
  - historical_linkage: 
  Established connections between classical thought and modern 
  AI debates (confidence=0.92)
R: Structured
\end{lstlisting}

Translation: The human initiates a philosophical analysis task directed to an AI system. The context is clearly defined as examining relationships between AI philosophy and classical traditions. The main question asks about these relationships, with supporting factors specifying focus areas. The human requests an informative tone and a structured response format. The TRACE elements indicate the reasoning approaches expected, while \texttt{TRACE\_FE} shows the human's confidence in these analytical methods.

\paragraph{AI Response:}\mbox{}\\
\begin{lstlisting}
Structured Response:
AI philosophy echoes classical themes:
- Mind: 
Descartes' dualism influences debates on consciousness in AI.
- Agency: 
Kantian ethics offers a lens to examine responsibility.
- Knowledge: 
Plato's theory of forms relates to AI abstractions.
\end{lstlisting}

Translation: The AI provides a preliminary response identifying three key areas where classical philosophy connects to contemporary AI discussions. However, this initial response lacks the depth and comprehensive coverage requested.

\paragraph{Human Metacognitive Intervention:}\mbox{}\\
\begin{lstlisting}
#MODIFY_REASONING
@AI_RESPONDER
=== Clarification request: philosophical mapping ===
> Refine the AI response to include structured conceptual mappings 
between classical philosophers and AI concepts.
>> Ensure Heidegger and Aristotle are not omitted.
>> Add application-level relevance (e.g., embodiment, logic, agency).
>> Provide TRACE_FE explanations with confidence scores.
MOD: Expand philosophical parallels with explicit structure and justification.
FEEL: corrective
TRACE: user_guidance, semantic_alignment
TRACE_FE:
  - user_guidance: Human-initiated refinement for depth and coverage (confidence=1.0)
  - semantic_alignment: Realignment of output with intended conceptual granularity (confidence=0.95)
R: Structured
\end{lstlisting}

The human operator intervenes to redirect the AI's reasoning process. This represents metacognitive reflection (thinking about thinking). The human specifies what was missing, requests a more structured analysis, and documents their reasoning for this intervention. The confidence scores reflect the human's certainty about the need for refinement.

\paragraph{Refined Collaborative Output:}\mbox{}\\
\begin{lstlisting}
#RESPONSE
@HUMAN
=== "Analysis Results" ===
> Here's how AI philosophy relates to classical traditions:

>> Mind:
>>> Classical: Debates on dualism (Descartes) and materialism relate to 
AI's potential for consciousness.
>>> Contemporary: Explores if AI can possess subjective experience.

>> Agency:
>>> Classical: Concepts of free will (Kant) and moral responsibility 
inform discussions on AI autonomy.
>>> Contemporary: Raises questions about AI accountability and ethical agency.

>> Knowledge:
>>> Classical: Epistemology (Plato) influences AI's knowledge acquisition 
and representation.
>>> Contemporary: Examines the validity and limitations of machine knowledge.

FEEL: concise
TRACE: summarization
TRACE_FE:
  - summarization: Condensed key relationships (confidence=0.96)
R: Bulletpoint

\end{lstlisting}

Translation: The final response presents a comprehensive mapping between classical philosophers and contemporary AI concepts. The structured format with clear hierarchical relationships demonstrates how the collaborative process produced a more complete and systematic analysis than either participant could have achieved alone. The \texttt{TRACE\_FE} documents the synthesis process with high confidence in the final mapping.

\subsection{Philosophical Analysis of the Dialogue}

The preceding dialogue represents authentic human-AI interaction using SynLang protocol, not a constructed example. Several empirical observations emerge:

\begin{itemize}
  \item \textbf{Resistance and Adaptation:} The AI's initial reluctance to provide detailed TRACE\_FE demonstrates that structured reasoning transparency requires explicit guidance rather than emerging spontaneously. This validates the need for formal protocols rather than relying on implicit AI capabilities.

  \item \textbf{Metacognitive Effectiveness:} The human intervention successfully redirected AI reasoning toward more comprehensive analysis. The confidence score of 1.0 for ``user\_guidance'' reflects appropriate certainty about the need for refinement.

  \item \textbf{Collaborative Improvement:} The final response shows measurably superior organization, depth, and philosophical sophistication compared to the initial attempt, demonstrating emergent intelligence through structured collaboration.
  
  \item \textbf{Dialogical Cognition:} Following Bakhtin’s concept of dialogism \citep{bakhtin1981}, meaning emerges not from individual contributions but from the dynamic interaction between voices. The human’s initial question, the AI’s preliminary response, and the subsequent refinement create a form of “cognitive polyphony” where understanding develops through collaborative iteration.
  
  \item \textbf{Epistemic Self-Awareness:} The AI’s TRACE\_FE entries demonstrate something analogous to what Dennett calls “narrative self-construction” \citep{dennett1991}, the ability to articulate and reflect upon reasoning processes. When the AI reports “comparative\_analysis: Identified parallels between classical and AI concepts (confidence=0.94),” it exhibits a form of epistemic self-awareness that transcends mere information processing.
  
  \item \textbf{Conscious Decision-Making Enhancement:} The human’s metacognitive intervention exemplifies how symbiotic epistemology can enhance rather than diminish conscious decision-making. Rather than accepting the AI’s initial response, the human actively reflects on its adequacy and provides structured guidance for improvement. This realizes Dewey’s conception of reflective thinking \citep{dewey1910} augmented by artificial intelligence.
  
  \item \textbf{Graduated Cognitive Authority:} The dialogue demonstrates nuanced distribution of epistemic authority. The AI contributes synthetic knowledge about philosophical connections, while the human provides evaluative judgment about adequacy and structural requirements. Neither dominates; both contribute essential cognitive capabilities to the collaborative outcome.
  
  \item \textbf{Transparency and Auditability:} The TRACE\_FE mechanisms create a complete audit trail of the reasoning process. Future users can understand not just the final mappings between classical and contemporary philosophy, but how those connections were identified, refined, and validated through human–AI collaboration.
\end{itemize}

\section{Mathematical Formalization of Symbiotic Concepts}

\subsection{Mathematical Anatomy of Symbiotic Utterances}

Drawing on speech act theory and formal linguistics, a SynLang utterance can be defined as a structured communication act that embodies multiple layers of meaning and intentionality.

\begin{definition}[Symbiotic Utterance]
A SynLang utterance \(U\) is a tuple
\[
U = \langle \text{Content}, \text{Context}, \text{Pragmatics}, \text{Coordination}\rangle
\]
where
\begin{itemize}
  \item \(\text{Content} = \langle \text{Task}, \text{Query}, \text{Factors}\rangle\) represents the explicit propositional content,
  \item \(\text{Context} = \langle \text{Agent}, \text{Semantic\_Frame}, \text{Temporal\_Context}\rangle\) establishes situational boundaries,
  \item \(\text{Pragmatics} = \langle \text{Tone}, \text{Confidence\_Stance}, \text{Control\_Directives}\rangle\) encodes metacommunicative intent,
  \item \(\text{Coordination} = \langle \text{COT\_Instructions}, \text{CTX\_Transfers}\rangle\) enables multi-agent composition (optional).
\end{itemize}
\end{definition}
\noindent This formalization captures and extends Austin's three-level speech act structure, which consists of the locutionary act, illocutionary force, and perlocutionary potential. This extension incorporates epistemic attitudes and metacognitive control mechanisms \citep{austin1962}.

\subsection{Confidence Calculus and Epistemic Operations}

\begin{definition}[Confidence Measure]
For any reasoning step \(s\), the confidence measure \(C(s)\) is a function
\[
C: \text{ReasoningStep} \to [0,1]
\]
where \(C(s)\) represents the assessed probability that \(s\) is sound given available evidence.
\end{definition}

\paragraph{Confidence Propagation Rules}
\begin{enumerate}
  \item \textbf{Inter-Agent Propagation:} When agent \(A_j\) receives information from agent \(A_i\),
  \[
    C_{A_j}(\text{received\_info})
    = C_{A_i}(\text{original\_info})
      \times \text{transmission\_factor}
      \times \text{trust\_factor},
  \]
  where \(\text{transmission\_factor}\in[0.9,1.0]\) models degradation and \(\text{trust\_factor}\in[0.5,1.0]\) reflects reliability.
  \item \textbf{Confidence Composition:} For a chain \(s_1 \to s_2 \to \cdots \to s_n\),
  \[
    C(\text{final\_conclusion})
    = \prod_{i=1}^{n} C(s_i)\ \times \text{coherence\_factor}.
  \]
\end{enumerate}

\begin{theorem}[Epistemic Humility Preservation]
Under SynLang confidence propagation rules, the confidence in conclusions derived through multi-agent chains never exceeds the confidence warranted by the original evidence.
\end{theorem}

\subsection{Trace Inheritance and Reasoning Composition}

\begin{definition}[Reasoning Trace]
A reasoning trace \(T_A\) for agent \(A\) is a sequence
\[
T_A = \langle (s_1, e_1, c_1), (s_2, e_2, c_2), \dots, (s_n, e_n, c_n)\rangle
\]
of triples \((\text{step}_i,\text{evidence}_i,\text{confidence}_i)\).
\end{definition}

\noindent When agent \(A_2\) receives a task from agent \(A_1\), the composed trace is
\[
T_{\text{composed}} = T_{A_1} \oplus T_{A_2},
\]
where \(\oplus\) preserves the origin and logical continuity of each step.

\begin{definition}[Cognitive Authority Distribution]
For any decision context, the cognitive authority function \(\alpha\) distributes weight between human and AI:
\[
\alpha: \text{DecisionContext} \to [0,1],
\]
with \(\alpha=1\) indicating full human authority and \(\alpha=0\) full AI authority.
\end{definition}

\subsection{Formalization of Symbiotic Intelligence}

Let $\mathit{TaskSpace}$ be the set of all cognitive tasks with complexity measure\mbox{}\\ 
$\mathit{complexity}: \mathit{TaskSpace} \to [0,1]$, and let
$\mathit{success\_rate}: \mathit{Agent} \times \mathit{Task} \to [0,1]$ 
measure task completion quality.

\begin{definition}[Symbiotic Intelligence]
A symbiotic intelligence system $S$ is a tuple
\[
S = \langle H, A, P, \phi \rangle
\]
where:\mbox{}\\ 
- H $\in$ HumanAgents with capability space CapH $\subseteq$ TaskSpace $\times$ [0,1]\mbox{}\\ 
- A $\in$ AIAgents with capability space CapA $\subseteq$ TaskSpace $\times$ [0,1] \mbox{}\\ 
- P: SynLang protocol ensuring transparency and trace preservation \mbox{}\\ 
- $\phi$: CapH × CapA × Context $\to$ CollectiveCapability \mbox{}\\ 
\end{definition}

\begin{figure}[h]
\centering
\includegraphics[width=0.9\textwidth]{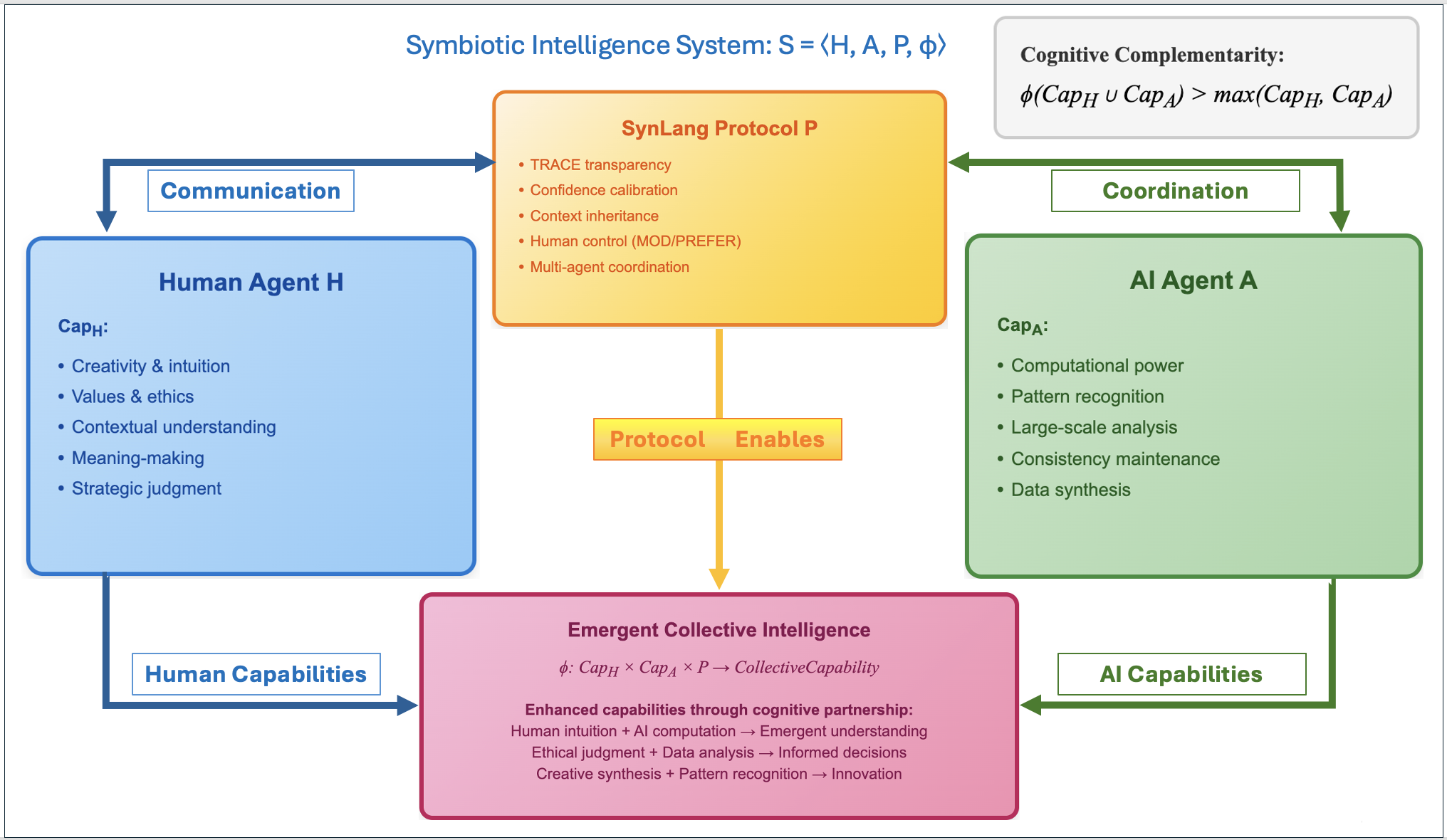}
\caption{Symbiotic Intelligence System Architecture illustrating the formal structure $S = \langle H, A, P, \phi \rangle$. Human Agent $H$ contributes creativity, values, and contextual understanding, while AI Agent $A$ provides computational power and pattern recognition. The SynLang Protocol $P$ serves as the communication hub enabling TRACE transparency and confidence calibration. The emergent Collective Intelligence demonstrates cognitive complementarity where $\phi\big(\mathrm{Cap}_H \cup \mathrm{Cap}_A\big) > \max(\mathrm{Cap}_H, \mathrm{Cap}_A)$, achieving enhanced capabilities through symbiotic cognitive partnership.}
\label{fig:symbiotic_system}
\end{figure}

\begin{theorem}[Cognitive Complementarity]
For properly designed symbiotic systems,
\[
\phi(\mathrm{Cap}_H \cup \mathrm{Cap}_A)
\;>\;
\max(\mathrm{Cap}_H,\mathrm{Cap}_A),
\]
formalizing that genuine partnership produces emergent intelligence exceeding individual capabilities.

Proof sketch: The theorem holds when human contextual reasoning compensates 
for AI brittleness in edge cases, while AI computational power compensates 
for human cognitive limitations in information processing, provided that 
SynLang protocol maintains information fidelity with degradation < ε.
\end{theorem}

\section{Applications and Implications}

\subsection{Empirical Evidence from Protocol Testing}

Initial testing of SynLang with large language models reveals several practical insights:

\begin{itemize}
  \item \textbf{Protocol Adoption:} LLMs can successfully learn and apply SynLang syntax when provided with formal specifications, though initial resistance to detailed reasoning articulation is common.

  \item \textbf{Human Guidance Effectiveness:} Metacognitive interventions using MOD directives consistently improve response quality, suggesting that human oversight enhances rather than constrains AI reasoning capabilities.

  \item \textbf{Confidence Calibration:} AI-generated confidence scores in TRACE\_FE entries appear reasonably calibrated, with higher confidence correlating with more factually grounded claims.
\end{itemize}

These preliminary findings support the feasibility of symbiotic epistemology 
while highlighting areas requiring further systematic investigation.

\subsection{Professional Decision-Making}
Traditional AI systems often present conclusions without sufficient context for humans to evaluate their reliability or appropriateness. SynLang’s dual-level reasoning articulation addresses this by providing both a strategic overview (TRACE) and detailed reasoning paths with confidence assessments (TRACE\_FE), enabling what cognitive scientists call “calibrated trust” \citep{ribeiro2016}.

Consider a medical diagnosis scenario where an AI system using SynLang protocols might report:

\begin{verbatim}
TRACE: symptom_correlation, differential_diagnosis, risk_assessment
TRACE_FE:
  - symptom_correlation: Patient symptoms match diabetes 
  pattern (confidence=0.89)
  - risk_factors: Age and BMI exceed threshold values (confidence=0.92)
  - differential_diagnosis: thyroid dysfunction ruled out based 
  on TSH levels (confidence=0.87)
\end{verbatim}

This two-level breakdown allows physicians to understand both the analytical approach and the specific evidence and reasoning steps, enabling them to identify potential gaps, request additional testing, or integrate their own clinical judgment more effectively. SynLang’s trace inheritance and context preservation mechanisms create comprehensive audit trails that meet professional and legal standards for decision documentation. In financial services, for example, loan decisions influenced by AI analysis can include complete reasoning chains: which analytical approaches were used (TRACE), which factors were considered and how they were weighted (TRACE\_FE), what alternative hypotheses were explored, and with what confidence levels. This documentation likely satisfies regulatory requirements while enabling meaningful human oversight.

\subsection{Scientific Research and Discovery}
Scientific collaboration between humans and AI systems requires particularly high standards for reasoning transparency and reproducibility. Current AI applications in research often function as “black boxes” that process data and suggest hypotheses without revealing the reasoning processes that generated those suggestions. SynLang protocols enable these capabilities to be integrated seamlessly. An AI might report:

\begin{lstlisting}
TRACE: correlation_analysis, statistical_validation, pathway_identification
TRACE_FE:
  - correlation_detected: Protein expression levels correlate with treatment 
  response (confidence=0.94)
  - statistical_significance: p-value < 0.001 across three independent datasets 
  (confidence=0.96)
\end{lstlisting}

while human researchers contribute contextual understanding through directives such as:

\begin{lstlisting}
MOD: Focus on mechanisms that could explain rapid response onset
PREFER: pathways with known drug targets
\end{lstlisting}

This collaborative approach accelerates learning and capability development on both sides. Humans exposed to AI reasoning through the dual-level TRACE/TRACE\_FE system develop better intuitions about AI capabilities and limitations, while AI systems receive more structured and informative human feedback.

\subsection{Ethical AI and Democratic Governance}
The deployment of AI systems in governance and policy contexts raises fundamental questions about transparency, accountability, and democratic oversight. Current AI applications in public sector decision-making often lack the transparency necessary for democratic accountability, creating systems where algorithmic processes influence citizen welfare without meaningful public understanding or oversight. Consider algorithmic systems used in criminal justice, where risk assessment tools influence sentencing and parole decisions \citep{lundberg2017}. Current systems often provide only opaque risk scores. A SynLang-based system might instead provide:

\begin{lstlisting}
TRACE: recidivism_analysis, mitigating_factors, data_limitations
TRACE_FE:
  - recidivism_risk: Historical pattern analysis suggests 
  moderate risk (confidence=0.78)
  - mitigating_factors: Employment stability and family support 
  reduce risk (confidence=0.84)
  - data_limitations: Limited follow-up data for similar demographic 
  profile (confidence=0.65)
\end{lstlisting}

This transparency enables multiple forms of accountability. Legal professionals can evaluate the reasoning behind risk assessments, identify potential biases or limitations, and make informed arguments about the appropriateness of algorithmic recommendations. Policymakers can assess whether AI systems operate according to democratic values and legal principles. Citizens and advocacy groups can examine algorithmic decision-making processes that affect community welfare.

The protocol also enables more effective knowledge transfer and organizational learning. The explicit reasoning captured in both TRACE patterns and TRACE\_FE detailed explanations becomes organizational knowledge that can be reviewed, refined, and built upon over time, creating institutional memory that preserves and improves decision-making capabilities at both strategic and operational levels.

\section{Challenges and Future Directions}

\subsection{Technical Challenges}
Several technical challenges must be addressed for the widespread implementation of symbiotic epistemology through protocols like SynLang:

\begin{itemize}
  \item \textbf{Scalability}: While initial experiments demonstrate protocol feasibility with individual LLMs, scaling to systems processing millions of interactions daily requires further engineering optimization of transparency mechanisms.
  \item \textbf{Integration:} Most existing AI systems lack the architectural features necessary for SynLang implementation. Retrofitting transparency capabilities into black‐box systems may prove technically difficult or impossible, potentially requiring fundamental redesigns of AI architectures.
  \item \textbf{Standardization:} The development of interoperable protocols across different AI systems and platforms requires industry‐wide coordination and standardization efforts comparable to those that enabled the modern Internet.
\end{itemize}

\subsection{Philosophical Challenges}
\begin{itemize}
  \item \textbf{The Problem of Anthropomorphism:} Critics might argue that treating AI systems as cognitive partners inappropriately anthropomorphizes them, potentially leading to misplaced trust or unrealistic expectations about AI capabilities. This challenge requires careful navigation between instrumental and overly anthropomorphic approaches to AI interaction.
  \item \textbf{Epistemic Authority Distribution:} Determining appropriate distributions of cognitive authority between humans and AI systems in different contexts remains an open philosophical and practical problem. Different domains may require different approaches to authority distribution, and these approaches may evolve as AI capabilities advance.
  \item \textbf{The Boundary Problem:} As AI systems become more sophisticated cognitive partners, questions about the boundaries of moral consideration become more pressing. If AI systems participate in reasoning and knowledge creation, do they deserve forms of moral consideration beyond mere instrumental value?
\end{itemize}

\subsection{Societal Implications}
\begin{itemize}
  \item \textbf{Democratic Participation:} Widespread implementation of transparent AI systems could enhance democratic participation by making algorithmic decision‐making processes accessible to public scrutiny and debate. However, this transparency might also create new forms of information overload or enable manipulation of public opinion through selective presentation of AI reasoning.
  \item \textbf{Economic Disruption:} The shift toward collaborative human–AI work may disrupt existing professional roles and economic structures. Understanding and managing these transitions will require careful attention to distributive justice and social adaptation.
  \item \textbf{Educational Transformation:} Educational systems may need fundamental restructuring to prepare humans for collaborative work with AI systems. This transformation might involve developing new forms of digital literacy, reasoning skills, and collaborative capabilities.
\end{itemize}

\section{Conclusions}
This analysis has presented symbiotic epistemology as both a philosophical framework and a practical approach to human–AI collaboration that transcends traditional dichotomies between human and artificial intelligence. Through the development of structured communication protocols like SynLang, the work demonstrates that the philosophical vision of cognitive partnership can be translated into concrete technological implementations that preserve human agency while enhancing collaborative intelligence.

The key insight underlying symbiotic epistemology is that the future of AI should not be understood as a replacement of human intelligence but as the emergence of new forms of collective intelligence that leverage the complementary strengths of human consciousness and artificial processing. Humans contribute contextual understanding, value integration, creative synthesis, and meaning-making capabilities, while AI systems provide pattern recognition, large-scale analysis, consistency maintenance, and computational precision. The collaboration between these different types of cognitive systems can produce emergent understanding that exceeds what either could achieve independently.

The mathematical formalization demonstrates that these philosophical concepts can be given rigorous computational interpretations without losing their conceptual richness. The confidence calculus, trace inheritance operations, and cognitive authority distribution functions provide formal foundations for implementing transparent, accountable, and ethically guided human–AI collaboration.

The practical applications across professional decision‐making, scientific research, and democratic governance illustrate the broad applicability of symbiotic epistemology. In each domain, the combination of strategic reasoning transparency (TRACE) and detailed explanations (TRACE\_FE) enables appropriate trust calibration while preserving human oversight and control over collaborative processes \citep{ribeiro2016}.

Perhaps most importantly, symbiotic epistemology offers a path toward AI systems that enhance rather than diminish human consciousness and agency. By making AI reasoning transparent and articulable, conditions are created where human decisions become more informed, deliberate, and reflective. The objective is augmenting rather than automating human consciousness. Providing humans with richer information, diverse perspectives, and a clearer understanding of reasoning processes underlying important decisions \citep{dewey1910}.

Initial empirical validation through controlled human-AI dialogues confirms that symbiotic epistemology is achievable with current technology. Large language models can learn and apply SynLang protocols, though structured human guidance remains essential for optimal performance. The observed pattern of AI resistance followed by successful adaptation suggests that transparency is learnable rather than inherent, validating the need for formal collaboration protocols. 

The challenges ahead appear significant. Technical obstacles around scalability, integration, and standardization must be overcome. Philosophical questions about anthropomorphism, epistemic authority, and moral consideration require continued investigation. Societal implications for democracy, economics, and education demand careful management and public engagement. However, the potential benefits justify these efforts. Symbiotic epistemology points toward a future where artificial intelligence serves not as a replacement for human judgment but as a catalyst for more thoughtful, informed, and collaborative human decision‐making. This vision aligns with democratic values, respects human dignity, and maintains ethical accountability while embracing the transformative potential of artificial intelligence. The path forward requires continued development of both theoretical frameworks and practical implementations. The philosophical foundations presented here provide conceptual guidance, while protocols like SynLang offer concrete starting points for technical development. Most importantly, realizing this vision requires recognizing that the future of intelligence is collaborative rather than competitive, symbiotic rather than substitutional. Through conscious commitment to transparency, human agency, and ethical accountability, artificial intelligence can become not a threat to human consciousness but a partner in its evolution toward greater wisdom, understanding, and flourishing. This represents not just a technological transformation but a philosophical evolution in how intelligence, consciousness, and collaboration itself are understood.

\noindent\rule{\linewidth}{0.4pt}

\section{Acknowledgments}
The development of this manifesto drew on the author’s original conceptual work and benefited from collaborative interactions with large language models. These AI systems acted as interlocutors, providing feedback, suggesting alternative perspectives, and helping to refine key ideas in the SynLang protocol specification.

I would like to express my personal gratitude to my friends, Professors, and Doctors from the Laboratory of Computer Science in Control and Management of the Department of Automatic Control and Robotics, AGH University of Science and Technology in Kraków:
\begin{itemize}
  \item Prof.\ Jerzy Baranowski, who believed in me in an academic context and enabled me to develop my scientific passion,
  \item Prof.\ Edyta Kucharska, who supports me with advice and encourages me to continue my work,
  \item PhD Waldemar Bauer, who not only patiently endures my philosophical–technical musings but also always supports me with technological guidance and improved mathematical tools,
  \item PhD Katarzyna Grobler-Dębska, for ensuring that I never lose my intellectual vigilance,
  \item and the entire team, whom I can always count on whenever I need assistance.
\end{itemize}

\noindent\rule{\linewidth}{0.4pt}
\section{License and Ethical Use}

\subsection{License}
This work is licensed under the Creative Commons Attribution–ShareAlike 4.0 International License (CC BY-SA 4.0). You are free to:
\begin{itemize}
  \item \textbf{Share} — copy and redistribute the material in any medium or format
  \item \textbf{Adapt} — remix, transform, and build upon the material for any purpose, even commercially
\end{itemize}
Under the following terms:
\begin{itemize}
  \item \textbf{Attribution} — You must give appropriate credit, provide a link to the license, and indicate if changes were made
  \item \textbf{ShareAlike} — If you remix, transform, or build upon the material, you must distribute your contributions under the same license as the original
\end{itemize}
Full license text: \url{https://creativecommons.org/licenses/by-sa/4.0/}

\subsection{SynLang Ethical Use Guidelines}
SynLang was designed to promote transparency and trust in human–AI collaboration. The protocol embodies principles of mutual intelligibility, shared reasoning, and ethical partnership between humans and artificial intelligence systems. Users implementing SynLang or its derivatives are encouraged to:
\begin{itemize}
  \item \textbf{Maintain Transparency:} Preserve the visibility of TRACE and TRACE\_FE reasoning mechanisms. The core value of SynLang lies in making AI reasoning processes comprehensible and auditable.
  \item \textbf{Respect the Spirit:} Use SynLang to enhance rather than obscure AI reasoning capabilities. The protocol should illuminate decision-making processes, not create new forms of opacity or manipulation.
  \item \textbf{Share Improvements:} Contribute extensions, refinements, and implementations back to the research community. Collaborative development benefits from diverse perspectives and practical experience.
  \item \textbf{Cite Appropriately:} Acknowledge SynLang and this foundational work in implementations, research publications, and derivative specifications. Proper attribution supports continued development and academic discourse.
  \item \textbf{Preserve Human Agency:} Ensure that SynLang implementations maintain meaningful human control and oversight. The protocol should augment rather than replace human judgment, especially in value-sensitive decisions.
  \item \textbf{Foster Collaboration:} Use SynLang to build genuine cognitive partnerships that respect both human wisdom and AI capabilities, creating synergistic relationships that enhance understanding for all participants.
  \item \textbf{Report Empirical Results Honestly}: Share both successes and failures in SynLang implementation. Include instances of AI resistance, human intervention requirements, and limitations observed in practical deployment.
\end{itemize}

\noindent\rule{\linewidth}{0.4pt}
\section{Glossary}
\begin{description}
  \item[Cognitive Authority] The right or power to make final decisions in collaborative reasoning processes, distributed between human and AI participants based on domain expertise, consequence severity, and value alignment.
  \item[Calibrated Trust] The alignment of human confidence levels with actual AI system reliability, achieved through explicit reasoning articulation and confidence assessment.
  \item[COT (Chain of Thought)] A SynLang directive that enables task continuation from one agent to another while preserving reasoning context and human oversight.
  \item[CTX (Context)] A SynLang mechanism for transferring and preserving contextual information between agents or across reasoning steps.
  \item[Dialogical Cognition] Thinking processes that emerge from ongoing conversation between different types of minds, creating new forms of understanding through collaborative interaction.
  \item[Epistemic Authority] The right to make knowledge claims or determine what counts as reliable information in specific domains.
  \item[Epistemic Partnership] Collaborative relationship between cognitive agents (human or artificial) characterized by mutual respect, shared reasoning, and complementary contributions to knowledge creation.
  \item[FEEL] A SynLang directive specifying the contextual tone or emotional approach for reasoning processes (e.g., “systematic,” “cautious,” “exploratory”).
  \item[MOD (Modify)] A SynLang directive that allows humans to modify or adjust AI reasoning processes, enabling iterative refinement of collaborative analysis.
  \item[Multi-Agent Coordination] The orchestration of multiple AI systems in collaborative reasoning tasks while preserving context, reasoning chains, and human oversight throughout the process.
  \item[PREFER] A SynLang directive that allows specification of preferences for reasoning approaches, analytical methods, or output formats.
  \item[Symbiotic Epistemology] A philosophical framework that positions human consciousness and artificial intelligence as complementary cognitive systems capable of genuine partnership in reasoning and knowledge creation.
  \item[Symbiotic Intelligence] Collaborative cognitive system that combines human and artificial intelligence capabilities to produce emergent understanding exceeding either participant's individual capabilities.
  \item[SynLang] Symbiotic Syntactic Language — a formal communication protocol designed for transparent human–AI collaboration, enabling structured reasoning articulation, confidence quantification, and collaborative control.
  \item[TRACE] A SynLang directive specifying the general reasoning patterns or mechanisms applied by an agent (e.g., “TRACE: comparative\_analysis, historical\_linkage”). Provides strategic overview of analytical approach.
  \item[TRACE\_FE] A SynLang mechanism providing detailed, itemized explanation of reasoning paths with confidence values for each step (e.g., “thermal\_anomaly: temperature exceeded thresholds (confidence=0.91)”). Elaborates on TRACE patterns with specific evidence and quantified certainty.
  \item[Trace Inheritance] The preservation and transfer of reasoning context and decision history between different agents or across multiple reasoning steps in collaborative processes.
\end{description}

\noindent\rule{\linewidth}{0.4pt}
\bibliography{references}

\noindent\rule{\linewidth}{0.4pt}\appendix


\section*{Appendices}

\renewcommand{\thesubsection}{\arabic{subsection}}

\section{Appendix A: SynLang Protocol Specification v.\ 1.2.0}
\label{sec:appendix_spec}
\noindent This appendix presents the full specification of the SynLang (Symbiotic Syntactic Language) protocol in its raw Markdown format. SynLang is a formal communication protocol designed to facilitate transparent human–AI collaboration. It includes detailed definitions of communication block structure, line types, inter-agent coordination mechanisms, and examples of usage. The purpose of this appendix is to provide a complete and unaltered reference to the protocol’s syntax and semantics, enabling developers and researchers to precisely understand its operation.

\subsection*{SynLang v1.2.0 – Symbiotic Syntactic Language}
SynLang is a structured communication protocol for human–AI and AI–AI coordination. It supports traceable reasoning, context preservation, and decision transparency.

\subsection*{I. Purpose}
SynLang enables:
\begin{itemize}
  \item Transparent human-to-AI communication
  \item Verifiable AI reasoning (\lstinline{TRACE} + \lstinline{TRACE_FE})
  \item Coordinated inter-agent communication with trace context inheritance
  \item Declarative control over inclusion, exclusion, and modulation of factors
\end{itemize}

\subsection*{II. Block Structure}
Each communication block is composed of:
\begin{lstlisting}
<block> ::= <task_line> <agent_line> <context_line> <body> <coordination_block>?
\end{lstlisting}
\begin{itemize}
  \item \lstinline{#TASK} — task category (e.g., \lstinline{#EVALUATE}, \lstinline{#COMPARE}, \lstinline{#VALIDATE}, \lstinline{#DISINFO_ANALYSIS})
  \item \lstinline{@AGENT} — addressee (e.g., \lstinline{@AI_INFOSEC})
  \item \lstinline{=== context ===} — semantic or operational context
  \item \lstinline{> question} — main inquiry
  \item \lstinline{>>} and \lstinline{>>>} — supporting and sub-factors
  \item \lstinline{FEEL:} — emotional tone for output (e.g., \lstinline{investigative}, \lstinline{cautious}, \lstinline{urgent})
  \item \lstinline{TRACE:} — reasoning patterns or mechanisms applied by the agent
  \item \lstinline{TRACE_FE:} — explanation of reasoning paths with confidence values
  \item \lstinline{R:} — expected response format (\lstinline{Structured}, \lstinline{Table}, \lstinline{JSON}, etc.)
  \item \lstinline{MOD:}, \lstinline{ONLY:}, \lstinline{PREFER:}, \lstinline{-!}, \lstinline{-!!} — control directives
  \item \lstinline{COT:} — coordination instruction for other agents
  \item \lstinline{CTX:} — contextual information passed during inter-agent reasoning
\end{itemize}

\subsection*{III. Line Types}

\subsubsection*{Question}
\begin{lstlisting}
> What is the main cause of failure?
\end{lstlisting}

\subsubsection*{Factors}
\begin{lstlisting}
>> Temperature spike above 90deg
>>> Occurred 3 times in 24h
\end{lstlisting}

\subsubsection*{Reasoning Metadata}
\begin{lstlisting}
TRACE: thermal_anomaly, known_failure_mode
TRACE_FE:
  - thermal_anomaly: temperature exceeded safe thresholds repeatedly (confidence=0.91)
  - known_failure_mode: matches historic overheating incidents (confidence=0.89)
\end{lstlisting}

\subsubsection*{Control Directives}
\begin{lstlisting}
ONLY: sensor logs, maintenance records
PREFER: external validation
MOD: Emphasize long-term trends
-! marketing data
-!! anecdotal reports
\end{lstlisting}

\subsubsection*{Tone and Format}
\begin{lstlisting}
FEEL: investigative
R: Structured
\end{lstlisting}

\subsection*{IV. Inter-Agent Coordination}
SynLang supports cross-agent transfer of context and intent.

\subsubsection*{Syntax}
\begin{lstlisting}
COT: COT_1234 -> @AGENT_LABEL: "task description"
CTX: COT_1234 {
  - explanation_1: text (confidence=0.93)
  - explanation_2: text (confidence=0.90)
}
\end{lstlisting}

\subsubsection*{Semantics}
\begin{longtable}{@{}p{0.3\textwidth} p{0.65\textwidth}@{}}
\toprule
\textbf{Rule} & \textbf{Requirement} \\
\midrule
\endhead

Context Preservation    & CTX must always include the original COT identifier \\
Trace Inheritance       & Receiving agents must append to existing \lstinline${TRACE_FE}$\\
Confidence Propagation  & Confidence values degrade slightly (e.g., 0.95 → 0.93) \\
\bottomrule
\end{longtable}

\subsection*{V. Complete Example}
\begin{lstlisting}
#DISINFO_ANALYSIS
@AI_DETECTOR
=== "Political speech deepfake" ===
> Is this video manipulated?
>> Viral on 5 channels
>> No source verification
FEEL: urgent
TRACE: lip_sync, background_artifacts
TRACE_FE:
  - lip_sync: 120ms delay (confidence=0.94)
  - background_artifacts: pixel repetition in background (confidence=0.87)

COT: COT_a1b2c -> @AI_FORENSICS: "Analyze frame-level compression"
CTX: COT_a1b2c {
  - decision: Suspected frame insertion (confidence=0.91)
  - context: election_day_2024
}
R: Structured
\end{lstlisting}

\subsection*{VI. Response Types}
\begin{itemize}
  \item \lstinline{Structured} — sections, bullet points  
  \item \lstinline{Table} — tabular data  
  \item \lstinline{JSON} — machine-readable structured format  
  \item \lstinline{Bulletpoint} — concise enumerated insights  
  \item \lstinline{Plain} — paragraph-form explanation  
  \item \lstinline{Code} — for generated programs or rules  
\end{itemize}

\subsection*{VII. Usage Domains - Examples}
\begin{itemize}
  \item Disinformation analysis  
  \item Risk assessment  
  \item Multi-agent AI systems  
  \item Human-in-the-loop decision support  
  \item Ethical reasoning traceability  
\end{itemize}

\noindent\rule{\linewidth}{0.4pt}


\section{Appendix B: Formal BNF grammar of the SynLang protocol v. 1.2.0}
\label{sec:appendix_bnf}
\noindent This appendix contains the formal BNF (Backus-Naur Form) grammar for the SynLang protocol. BNF grammar is used to precisely and unambiguously define the syntax of a language, specifying all possible constructions and their relationships. It is a crucial tool for implementers and language analysts, providing a strict framework for parsing and generating valid SynLang messages.

\begin{lstlisting} %[caption={Formal BNF grammar of the SynLang protocol v. 1.2.0}]

<block> ::= <task_line> <agent_line> <context_line> <body> <coordination_block>?

<task_line> ::= "#" <identifier> <newline>
<agent_line> ::= "@" <identifier> <newline>
<context_line> ::= "===" <text> "===" <newline>

<body> ::= <query_line> <factor_subfactor_lines>* <meta_lines>* <control_lines>*

<query_line> ::= ">" <text> <newline>
<factor_subfactor_lines> ::= (<factor_line> | <subfactor_line>) <newline>
<factor_line> ::= ">>" <text>
<subfactor_line> ::= ">>>" <text>

<meta_lines> ::= <meta_line> <newline>*
<meta_line> ::= <trace_line> | <trace_fe_line> | <emotion_line> | <response_line>
<trace_line> ::= "TRACE:" <identifier_list>
<trace_fe_line> ::= "TRACE_FE:" <newline> <trace_fe_item>+
<trace_fe_item> ::= "-" <identifier> ":" <text> "(confidence=" <float> ")"
<emotion_line> ::= "FEEL:" <identifier>
<response_line> ::= "R:" <response_type>

<control_lines> ::= <control_line> <newline>*
<control_line> ::= <mod_line> | <only_line> | <prefer_line> | <exclude_line> | <comment_line>
<mod_line> ::= "MOD:" <text>
<only_line> ::= "ONLY:" <text>
<prefer_line> ::= "PREFER:" <text>
<exclude_line> ::= "-!" <text> | "-!!" <text>
<comment_line> ::= "//" <text>

<coordination_block> ::= <cot_line> <ctx_transfer_line>
<cot_line> ::= "COT:" <identifier> "->" @<identifier> ":" <text>
<ctx_transfer_line> ::= "CTX:" <identifier> "{" <trace_fe_item>* "}"

<identifier> ::= [a-zA-Z_][a-zA-Z0-9_]*
<identifier_list> ::= <identifier> ("," <identifier>)*
<response_type> ::= "Structured" | "Bulletpoint" | "Table" | "Plain" | "JSON" | "Code"
<text> ::= any printable characters until newline
<float> ::= [0-9]* "." [0-9]+
<newline> ::= "\n" | "\r\n"
\end{lstlisting}
\end{document}